\documentclass[conference]{IEEEtran}
\IEEEoverridecommandlockouts
\usepackage{cite}
\usepackage{amsmath,amssymb,amsfonts}
\usepackage{algorithmic}
\usepackage{graphicx}
\usepackage{textcomp}
\usepackage{xcolor}
\usepackage{epsfig,url,stfloats,latexsym}
\usepackage{multirow,array}
\usepackage{hhline,bm}
\usepackage{subcaption}
\usepackage{verbatim}
\usepackage{epsfig}
\usepackage[utf8]{inputenc}
\usepackage{caption}
\usepackage{algorithm}
\usepackage{array}
\usepackage{color}
\usepackage{multirow}
\usepackage{makecell}
\usepackage{bbding}
\usepackage{pifont}
\usepackage{wasysym}
\usepackage{multirow}

\def\BibTeX{{\rm B\kern-.05em{\sc i\kern-.025em b}\kern-.08em
    T\kern-.1667em\lower.7ex\hbox{E}\kern-.125emX}}
\begin{document}

\title{Securing IoT Communication using Physical Sensor Data - Graph Layer Security with Federated Multi-Agent Deep Reinforcement Learning \\
}

\author{\IEEEauthorblockN{Liang Wang, Zhuangkun Wei and Weisi Guo}	
	\IEEEauthorblockA{School of Aerospace, Transport and Manufacturing, Cranfield University, Milton Keynes, UK}
 
	\{liang.wang.133, zhuangkun.wei, weisi.guo\}@cranfield.ac.uk
}

\maketitle

\begin{abstract}
Internet-of-Things (IoT) devices are often used to transmit physical sensor data over digital wireless channels. Traditional Physical Layer Security (PLS)-based cryptography approaches rely on accurate channel estimation and information exchange for key generation, which irrevocably ties key quality with digital channel estimation quality. Recently, we proposed a new concept called Graph Layer Security (GLS), where digital keys are derived from physical sensor readings. The sensor readings between legitimate users are correlated through a common background infrastructure environment (e.g., a common water distribution network or electric grid). The challenge for GLS has been how to achieve distributed key generation. This paper presents a Federated multi-agent Deep reinforcement learning-assisted Distributed Key generation scheme (FD2K), which fully exploits the common features of physical dynamics to establish secret key between legitimate users. We present for the first time initial experimental results of GLS with federated learning, achieving considerable security performance in terms of key agreement rate (KAR), and key randomness.

\end{abstract}


\IEEEdisplaynontitleabstractindextext

\IEEEpeerreviewmaketitle

\section{Introduction}

Recently, with the advances in wireless sensor networks and embedded systems, our life has been revolutionized by a variety of Internet-of-Things (IoT) devices, which provide ubiquitous connections among environments. More and more IoT-based applications have been implemented in healthcare and smart cities. For instance, in a complex networked system, such as an oil, gas, or water network, the maintenance team can have 24/7 monitoring of the conditions of the network by deploying multiple embedded IoT devices. However, as the data obtained by IoT devices are normally related to privacy, sensitivity, and confidentially, which are easily captured by adversaries or eavesdroppers, the research challenge lies in how to secure the data through wireless communication.

\subsection{Physical Layer Security}

The traditional method for securing the data is to apply a complexity-based cryptography scheme, which, however, does not have an information theory-based secrecy guarantee, and is therefore vulnerable to post-quantum threat. In order to achieve the theoretical secure capacity, physical layer security (PLS)-based key generation has been viewed as an efficient technology, which extracts common randomness of features to generate secret keys between two legitimate users, i.e, Alice and Bob~\cite{256484}. Specifically, Alice and Bob keep measuring their common channel through a key generation protocol including channel probing, quantization, information reconciliation, and privacy amplification, and finally generate symmetric cipher keys at both ends~\cite{7809064}.
Current PLS-based key generation schemes leverage to extract common features from channel state information (CSI). In~\cite{10.1145/1409944.1409960}, the authors were the first to propose a sufficient key generation algorithm in WIFI system, in which both the received signal strength indicator (RSSI) and the peak of channel impulse response (CIR) were considered. In order to break the barrier that RSSI and the peak of CIR can only provide limited information from wireless channels, it was found that there are more CSI about multiple sub-carriers of orthogonal frequency-division multiplexing (OFDM)~\cite{6226870, halperin2011tool}. In~\cite{7063613}, Zhang~\emph{et al.} proposed a key generation scheme that exploits the subcarriers' channel response over time in OFDM system. In~\cite{6914340}, Xi~\emph{et al.} proposed a key generation scheme named KEEP, which utilized a validation recombination mechanism to generate keys from CSI measurements of all subcarriers. In~\cite{ruotsalainen2019experimental}, the authors conducted extensive experiments about the different configurations of long-range communication technology (LoRA) on the key generation performance. In~\cite{8519327}, Zhang~\emph{et al.} concluded that the movement of the LoRA devices has a significant influence on the received power in a large-scale environment and they proposed a differential value-based key generation scheme. Other key generation schemes have been extensively studied for various wireless scenarios, e.g., the fifth-generation (5G), and wireless body area network (WBAN). In~\cite{8433175}, the features of the virtual angle of arrival (AoA) and angle of departure (AoD) were utilized to generate keys in the millimetre wave multiple-input and multiple-output (mmWave MIMO) system and the authors analyzed the performance. In~\cite{zhang2021h2k}, Zhang~\emph{et al.} found that the interpulse intervals (IPIs) between two adjacent peaks of heartbeat signals can be a good random source for key generation in WBAN and they proposed a key establishment protocol, which utilized the electrocardiography (ECG) and photoplethysmography (PPG) signals to secure communications.

Although PLS-based key generation schemes have been widely applied in a variety of wireless networks, it is still not suitable for IoT-embedded networks. On the one hand, PLS assumes that the wireless channel between the legitimates must contain enough characteristics in terms of reciprocity, dynamics, and uniqueness~\cite{7944621}. That is to say, in order to establish reliable keys, IoT devices should acquire reasonable signal processing capability and rely on the high communication signal-to-noise ratio (SNR). On the other hand, most IoT devices are normally buried underground, which means that reasonable SNR can not be obtained. Therefore, PLS is not suitable when channel estimation is not reliable.

\subsection{Graph Layer Security}

In order to solve the issue that PLS requires a high SNR for channel estimation, Graph Layer Security (GLS)~\cite{s22103951} has become a promising approach. Generally, in networked systems, the embedded physical dynamics can be described by a common continuity equation, such as Navier Stokes for flow, and Nonlinear Schrodinger for optical transmission. The IoT devices deployed in networked systems can commonly detect the physical dynamics with high accuracy and they do not suffer from high SNR for channel estimation. However, the proposed scheme in~\cite{s22103951} utilized the overall physical dynamics among the networked system to generate the global keys for information encryption/decryption between legitimate users. The global keys could be easily captured by adversaries and the overall key generation system is not fully distributed. Motivated by the previous work in GLS and in order to make the generated keys robust against attacks from adversaries, we summarize the main contributions of this paper as follows:

\begin{itemize}
    \item We propose a FD2K scheme in which IoT devices can generate cipher keys based on their locally detected physical dynamics for information encryption and decryption.
    \item Considering the situation that physical dynamics of IoT devices stored locally could be easily detected by the adversary, we utilize the technique of federated learning during the training process to eliminate the process of data collection, therefore, the privacy and security of the physical dynamics are protected.
    \item The multi-agent deep reinforcement learning technique is applied to generate keys within the framework of centralized training and decentralized execution. Thus, the overall key generation system can be fully distributed when deploying.
\end{itemize}

The rest of the paper is organized as follows. The system model is described in Section~\ref{system_model}. Then, we introduce the proposed FD2K scheme in Section~\ref{proposed_scheme}. The simulation results are analyzed in Section~\ref{experiments}. Finally, the conclusion is given in Section~\ref{conclusion}.

\section{System Model}\label{system_model}

\begin{figure}[!htpb]
\centering
\includegraphics[scale=0.2]{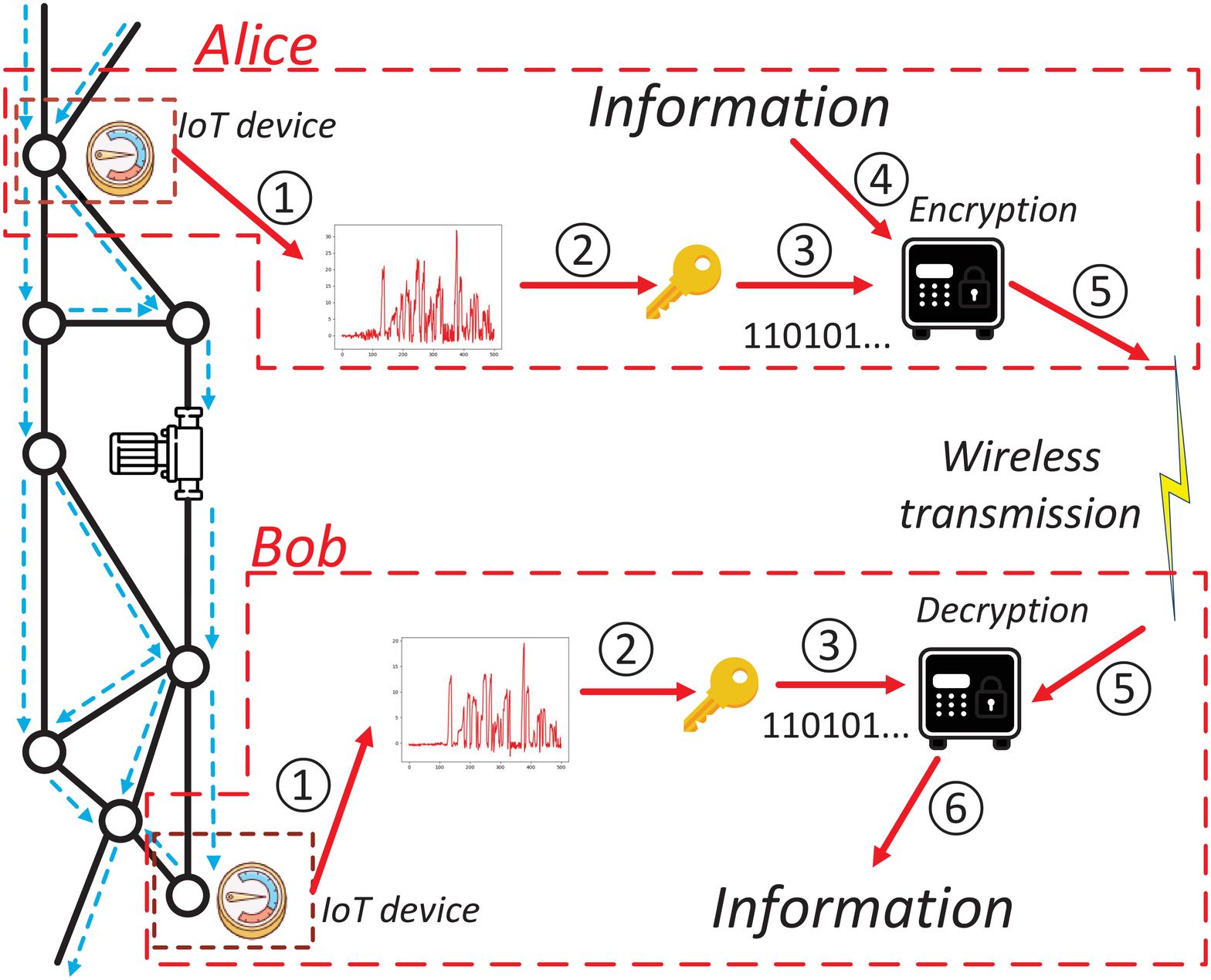}
\caption{Overall system model.}
\label{overall_system_model}
\end{figure}

We describe the system model in Fig.~\ref{overall_system_model}, where two legitimate users, namely Alice (A) and Bob (B), are placed to constantly monitor the physical signals of a networked system with equipped IoT devices. The key generation scheme consists of 6 stages, namely \textcircled{\small{1}} signal acquisition, \textcircled{\small{2}} common feature extraction, \textcircled{\small{3}} key generation, \textcircled{\small{4}} information encryption, \textcircled{\small{5}} wireless transmission, and \textcircled{\small{6}} information decryption. Note that the networked system can be a water, oil, or gas network and the physical signals can be pressure, flow rate, or temperature. For simplicity, in this paper, we consider the water distribution network as the networked system and water pressure as the physical signals.

We assume the overall process lasts $T$ time steps (TSs), in each of which, both Alice and Bob have $M$ detected physical values. Therefore, the physical signals detected in TS $t$ can be expressed as
\begin{equation}
    \begin{aligned}
        \bm{P^i_t} = [p^i_{t,1},...,p^i_{t,m}...,p^i_{t,M}],
    \end{aligned}
\end{equation}
where $i=\{A, B\}$, $p^i_{t,m}$ is the $m$-th signal value obtained by Alice or Bob in TS $t$.

Then, Alice and Bob can extract common features of obtained physical signals to generate cipher keys. Note that in this paper, we assume the key generation rate (KGR) is $1$, meaning the keys' length is $M$. We denote the keys generated in TS $t$ as  
\begin{equation}
    \begin{aligned}
        \bm{K^i_t} = [k^i_{t,1},...,k^i_{t,m}...,k^i_{t,M}],
    \end{aligned}
\end{equation}
where $i=\{A, B\}$, $k^i_{t,m} \in \{0, 1\}$ denotes the $m$-th binary bit of keys generated by Alice or Bob. 

After that, one legitimate user (e.g., Alice) can use his/her generated cipher keys to encrypt the information and then transmit the encrypted information through a wireless channel to another (e.g., Bob), with the same process for information decryption. As a result, secure communication between Alice and Bob is conducted.

The challenge of key generation in this system model lies in how to generate cipher keys in a distributed manner and protect the local physical signals simultaneously. More precisely, during the process of key generation, Alice and Bob have to fully exploit the common features of physical signals between each other without any information exchange. Besides, they have to generate well-qualified keys that satisfy the performance metrics, which we will introduce in Section~\ref{experiments}.

\section{Proposed FD2K Scheme}\label{proposed_scheme}

\begin{figure}[!htpb]
\centering
\includegraphics[scale=0.2]{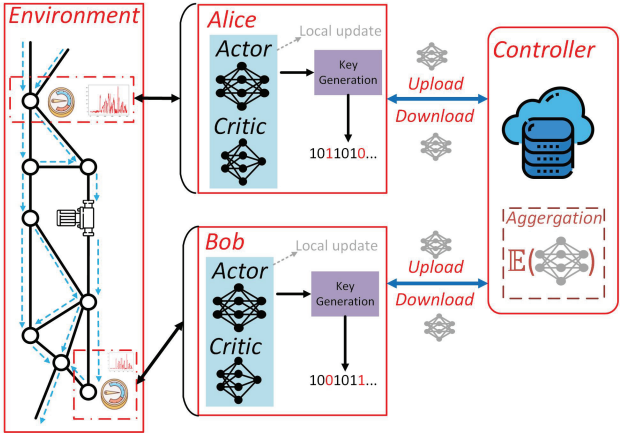}
\caption{Structure of proposed FD2K scheme.}
\label{proposed_scheme_diagram}
\end{figure}

In this section, we introduce our proposed FD2K scheme. First of all, the overall structure of FD2K is shown in Fig.~\ref{proposed_scheme_diagram}, which can be described within the framework of multi-agent deep reinforcement learning (MADRL) and is based on the partially observable Markov decision process (POMDP). Specifically, the networked system can be viewed as an environment. Alice and Bob are controlled by its dedicated agent, which equips with two deep neural networks (DNNs), namely actor $a^A_t=\pi^A(o^A_t|\theta^A_{\pi}), a^B_t=\pi^B(o^B_t|\theta^B_{\pi})$ for policy approximation and critic $Q^A(s_t, a_t|\theta^A_Q), Q^B(s_t, a_t|\theta^B_Q)$, where $t$ means the index of time step; $o^A_t, o^B_t$ denote the local observations of Alice and Bob; $a^A_t, a^B_t$ are the local actions generated actor networks, followed by policy functions $\pi^A(\cdot|\theta^A_{\pi}), \pi^B(\cdot|\theta^B_{\pi})$ with network parameters $\theta^A_{\pi}, \theta^B_{\pi}$; $s_t \triangleq \{o^A_t, o^B_t\}, a_t \triangleq  \{a^A_t, a^B_t\}$ are the state and action assembled by all agents; $Q^A(\cdot|\theta^A_Q), Q^B(\cdot|\theta^B_Q)$ are the action-value functions with network parameters $\theta^A_Q, \theta^B_Q$. Note that both actor and critic network has their own target network with the same structure, which can be expressed as $\pi^{A'}(\cdot|\theta^{A'}_{\pi}), \pi^{B'}(\cdot|\theta^{B'}_{\pi}), Q^{A'}(\cdot|\theta^{A'}_Q), Q^{B'}(\cdot|\theta^{B'}_Q)$ with parameters $\theta^{A'}_{\pi}, \theta^{B'}_{\pi}, \theta^{A'}_Q, \theta^{B'}_Q$ respectively. During the training process, the agents of Alice and Bob keep sending local observations and local actions to the environment. Then, they will receive the state, action and the environment will transfer to the next state. The agents learn to train the actor, critic networks from exploitation and exploration until the convergence is obtained, aiming at maximizing the accumulated reward $R_t=\sum_{t'=t}^{T}\gamma^{t'-t}r_{t'}$, where $\gamma \in [0, 1]$; $r_{t'}$ is the reward function in TS $t'$.

The training process of the traditional DRL-based system is implemented in a simulator that requires the established training data based on previously obtained physical signals, which have to be collected remotely and could be known by Eve. In order to protect the security and privacy of the local physical signals of Alice and Bob in the training process, federated learning (FL)~\cite{yang2019federated} is applied to eliminate the process of data collection. More precisely, in the framework of the proposed scheme, a controller is deployed in the proposed scheme. It first initializes a global actor network and downloads it to Alice and Bob. Then, Alice and Bob take local training for certain epochs and upload their local actor network models to the controller, which takes aggregation and obtains a new actor network. Both Alice and Bob repeat the process of downloading, local training, and uploading. The controller repeats aggregation until the network is adequately trained. Therefore, it is concluded that during the training process, the training data of agents are stored locally and only the network model is uploaded/downloaded, which can help to protect privacy and security.

\begin{algorithm}[htpb!]
	\scriptsize
	\caption{Proposed FD2K}\label{proposed_algorithm}
	\begin{algorithmic}[1]
        \STATE Controller initializes global actor network and downloads it to Alice $\pi^A(\cdot|\theta^A_{\pi})$ and Bob $\pi^B(\cdot|\theta^B_{\pi})$;\
        \STATE Initialize critic networks $Q^A(\cdot|\theta^A_Q), Q^B(\cdot|\theta^B_Q)$ with parameters $\theta^A_Q$, $\theta^B_Q$;\
        \STATE Initialize target networks: \\
        $\theta^{i'}_{\pi} \leftarrow \theta^i_{\pi}, i \in \{A, B\}$,\\
        $\theta^{i'}_Q \leftarrow \theta^i_Q, i \in \{A, B\}$;\\
        
        \STATE Initialize experience replay memory with the size of $N$;\

        \FOR{epoch $e=1$ to $e^{\textit{max}}$}
        \STATE Initialize $o^A_t$, $ o^B_t$;\
        \FOR{TS t=1 to T}
        \STATE Obtain $s_t \triangleq \{o^A_t, o^B_t\}$;\
        \STATE Obtain local actions $a^A_t$, $a^B_t$ according to:\ \\
        $a^A_t=\pi^A(o^A_t|\theta^A_{\pi})+\epsilon$ and $a^B_t=\pi^B(o^B_t|\theta^B_{\pi})+\epsilon$;\ \\
        \STATE Execute local actions $a^A_t$, $a^B_t$;\
        \STATE Obtain action $a_t \triangleq  \{a^A_t, a^B_t\}$;\
        \STATE Generate keys according to Algorithm~\ref{proposed_key_generation};\
        \STATE Calculate reward function $r_t$;\
        \STATE Transfer the environment to $s_{t+1} \triangleq \{o^A_{t+1}$, $o^B_{t+1}\}$;\
        \STATE Store transition $\{s_t, a_t, r_t, s_{t+1}\}$ into experience replay memory;\
        \IF{learning process starts}
        \STATE Random sample $n$ transitions from experience replay memory;\
        \STATE Update critic networks of Alice and Bob according to Eq.~\ref{loss_function};\
        \STATE Update actor networks of Alice and Bob according to Eq.~\ref{policy_gradient};\
        \STATE Update target networks with the rate $\rho$:\ \\
        $\theta^{i'}_{\pi} \leftarrow \rho \theta^i_{\pi} + (1-\rho)\theta^{i'}_{\pi}, i \in \{A, B\}$,\\
        $\theta^{i'}_Q \leftarrow \rho \theta^i_Q + (1-\rho)\theta^{i'}_Q, i \in \{A, B\}$;\\
        \ENDIF
        \ENDFOR
        \IF{$e$ mod $E==0$}
        \STATE Upload actor networks of Alice and Bob to the controller for federated updating;\
        \ENDIF
        \ENDFOR
	\end{algorithmic} 
\end{algorithm}

We further provide the overall pseudo code of the proposed FD2K scheme in Algorithm~1. First of all, the controller initializes a global actor network and downloads it to Alice and Bob. Then, Alice and Bob initialize their critic and target networks. In Line 4, in order to avoid divergence and achieve stable convergence performance, experience replay memory~\cite{mnih2015human} with the size of $N$ is deployed and initialized for storing the previous transitions. During the training process, Alice and Bob initialize their local observations in the first TS. Then, in each TS, their local observations are assembled to state $s_t$ and they obtain their local actions by actor networks. Note that in Line 9, an action noise $\epsilon$ is used for a better exploration and it decays with a fixed rate. After that, Alice and Bob take common features of physical signals to generate keys according to the obtained actions. More details about the process of key generation are explained in Algorithm \ref{proposed_key_generation}. Then, the reward function is calculated with the given generated keys of Alice and Bob in Eq.~\ref{reward_function}. From Line 14, the environment is transferred to the next state and the transition is stored in the experience replay memory. When the learning process starts, $n$ transitions are randomly sampled from the memory to train the networks. Specifically, the actor networks of Alice and Bob are trained by the following policy gradient method~\cite{10.5555/3009657.3009806}, whose equation can be described as
\begin{equation}\label{policy_gradient}
    \begin{aligned}
        \triangledown_{\theta^i_{\pi}}J \approx \mathbb{E}\big[\triangledown_{\theta^i_{\pi}}\pi^i\big(o^i_t|\theta^i_{\pi}\big)\triangledown_{a^i_t}Q^i(s_t,a_t|\theta^i_{Q})\big],
    \end{aligned}
\end{equation}
where $i=\{A,B\}$. The critic networks of Alice and Bod are trained by the loss function~\cite{lowe2017multi}, which is:
\begin{equation}\label{loss_function}
    \begin{aligned}
        L(\theta^i_Q) = \mathbb{E}\big[r_t+\gamma Q^{i'}(s_{t+1},a_{t+1}|\theta^{i'}_Q) - Q^i(s_t,a_t|\theta^i_Q)\big],
    \end{aligned}
\end{equation}
where $i=\{A,B\}$ and $a_{t+1}$ is the action of next state.

Then, the target networks are updated as well at a rate of $\rho$. Finally, from Line 23, both Alice and Bob upload their actor networks to the controller and the aggregation process is carried out to obtain a new network. In this paper, as the FedAvg~\cite{mcmahan2017communication} is applied, the aggregation process can be expressed as:
\begin{equation}
    \begin{aligned}
        \Bar{\theta_{\pi}}=\frac{\theta^A_{\pi}+\theta^B_{\pi}}{2},
    \end{aligned}
\end{equation}
where $\Bar{\theta_{\pi}}$ denotes the parameters of new network. 

Therefore, we define the local observations, and local actions of Alice and Bob as follows:
\begin{itemize}
    \item \textbf{Local observation}: we define the signal values detected by Alice and Bob in TS $t$ as their observations, which are $o^A_t = \bm{P^A_t}$ and $o^B_t = \bm{P^B_t}$.
    \item \textbf{Local action}: in our proposed FD2K scheme, the local actions are used to extract common features of the physical signals of Alice and Bob for key generation. Specifically, let us denote two vectors $\bm{A^A_t} \in \mathbb{R}^M, \bm{A^B_t} \in \mathbb{R}^M$ for Alice and Bob. For $m$-th values in $\bm{A^A_t}$ and $\bm{A^B_t}$, they have:
    \begin{equation}
        \begin{aligned}
            \bm{A^i_t}(m)=\begin{cases}
			1, & \text{if $a^i_t(m) \geq \lambda$},\\
            0, & \text{otherwise},
            \end{cases}
        \end{aligned}
    \end{equation}
    where $\lambda$ is the threshold, $i \in \{A, B\}$. Specifically, if $\bm{A^A_t}(m)=1$ or $\bm{A^B_t}(m)=1$, it means the $m$-th signal value of $\bm{P^A_t}$ or $\bm{P^B_t}$ has a common feature that can be used to generate keys.
    \item \textbf{Reward}: we define the reward function as follows:
    \begin{equation}\label{reward_function}
        \begin{aligned}
            r_t =& 1-\frac{\sum_{m=1}^{M}|\bm{K^A_t}(m)-\bm{K^B_t}(m)|}{M}\\&+\phi\frac{\sum_{m=1}^{M}\big(\bm{A^A_t}(m)+\bm{A^B_t}(m)\big)}{2M},
        \end{aligned}
    \end{equation}
    
    where $1-\frac{\sum_{m=1}^{M}|\bm{K^A_t}(m)-\bm{K^B_t}(m)|}{M}$ is the key agreement rate (KAR), which is one of the performance metrics of key generation. $\frac{\sum_{m=1}^{M}\big(\bm{A^A_t}(m)+\bm{A^B_t}(m)\big)}{2M}$ is the average percentage of common features applied to generate keys. $\phi$ is expressed as
    \begin{equation}
        \begin{aligned}
            \phi=\begin{cases}
			1, & \text{if $1-\frac{\sum_{m=1}^{M}|\bm{K^A_t}(m)-\bm{K^B_t}(m)|}{M} = 1$},\\
            0, & \text{otherwise},
            \end{cases}
        \end{aligned}
    \end{equation}

    Overall, the reward function is designed to guarantee that generated keys can satisfy the KAR and fully utilize the common features of physical signals.
    
\end{itemize}

\begin{algorithm}[!htpb]
	\scriptsize
	\caption{Key Generation}\label{proposed_key_generation}
	\begin{algorithmic}[1]
        \STATE Obtain $\bm{P^A_t}, \bm{P^B_t}, \bm{A^A_t}, \bm{A^B_t}$;\\
        \STATE Initialize $\bm{K^A_t},\bm{K^B_t}$;\\
        \FOR{m=1 : M}
        \FOR{$i \in \{A, B\}$}
        \IF{$\bm{A^i_t}(m)==1$ and $\bm{P^i_t}(m) \geq \bm{P^i_t}(m-1)$}
        \STATE $\bm{K^i_t}(m)\leftarrow 1$;\\
        \ELSE 
        \STATE $\bm{K^i_t}(m)\leftarrow 0$ ;\\
        \ENDIF
        \ENDFOR
        \ENDFOR
        \RETURN $\bm{K^A_t}$ and $\bm{K^B_t}$;\\
	\end{algorithmic} 
\end{algorithm}

We further give the overall algorithm design for key generation in Algorithm~\ref{proposed_key_generation}. Note that in this paper, the differentiation of physical signals is captured as the common feature. Specifically, $\bm{P^A_t}, \bm{P^B_t}, \bm{A^A_t}, \bm{A^B_t}$ are first obtained and $\bm{K^A_t}, \bm{K^B_t}$ are initialized. Then, the feature extraction for key generation is started. Specifically, for Alice, if $\bm{A^A_t}(m)==1$ and $\bm{P^A_t}(m)\geq \bm{P^A_t}(m-1)$, we have $\bm{K^A_t}(m)\leftarrow 1$, otherwise, we have $\bm{K^A_t}(m)\leftarrow 0$. Similarly, Bob has the same key generation scheme. Finally, the generated keys of Alice and Bob are obtained.

\section{Experimental Results}\label{experiments}

\begin{figure}[!htpb]
\centering
\includegraphics[scale=0.2]{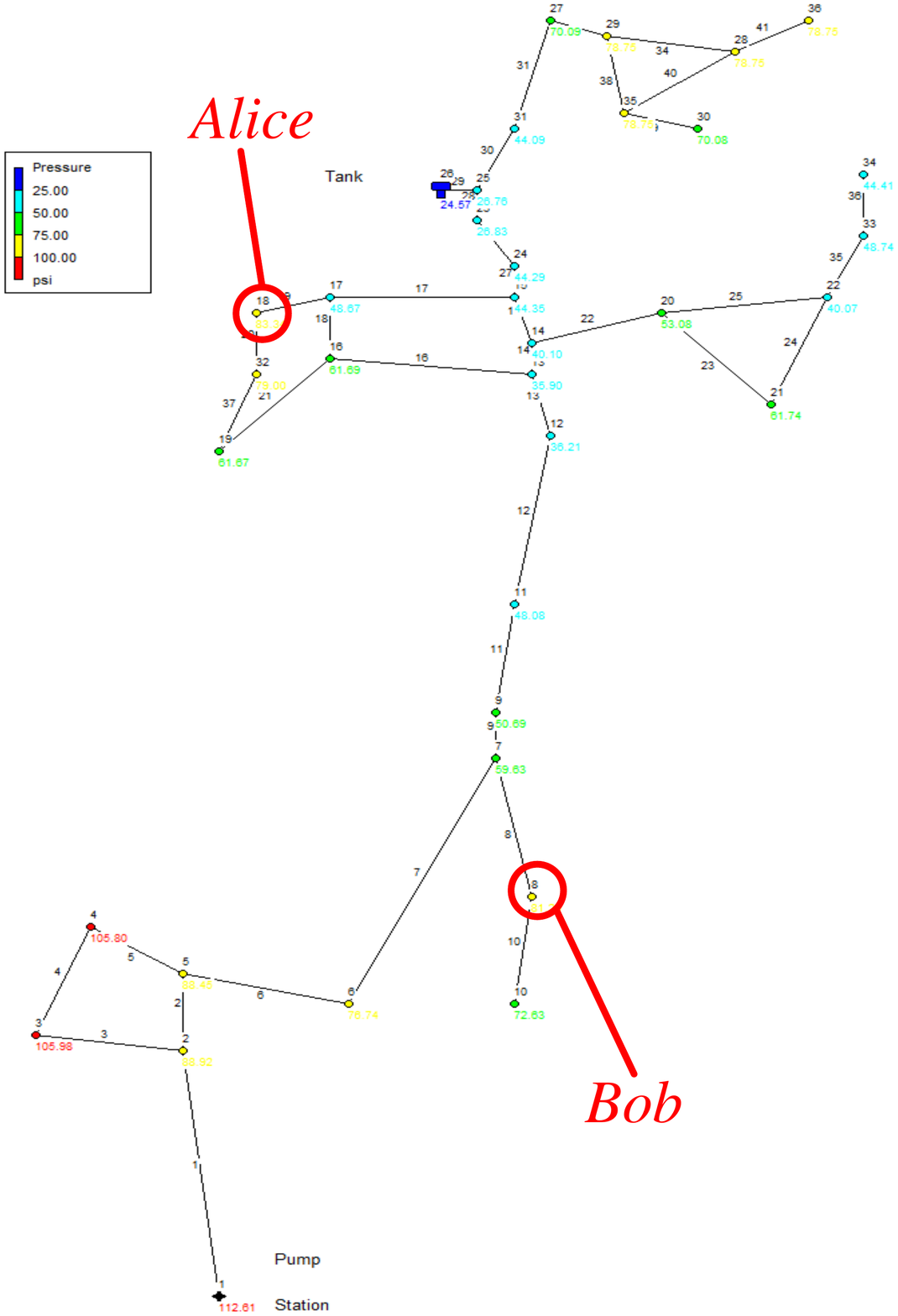}
\caption{The structure of the water distribution network.}
\label{water_network}
\end{figure}

\begin{figure}[!htpb]
\centering
\includegraphics[scale=0.4]{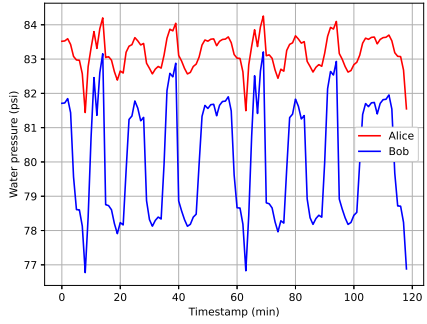}
\caption{The physical signals detected by Alice and Bob.}
\label{pressure_value}
\end{figure}

In this section, we conduct experiments to analyze the performance of the proposed FD2K scheme in a simulated water distribution network, which is designed by an open-sourced simulation software - EPANET~\cite{rossman2000epanet}. Specifically, it is based on a small region of the Cherry Hill/Brushy Plains distribution system and has a total demand of 0.90 million gallons per day (MGD), one tank and 6.8 miles of pipe. The overall structure is shown in Fig.~\ref{water_network}, in which we choose Node 18 and Node 8 as Alice and Bob. We define the water pressures of Alice and Bob as the physical signals and we record the previous 2 hours' water pressure for training. The snapshot of the water pressures of Alice and Bob is shown in Fig.~\ref{pressure_value}.

The proposed FD2K algorithm is implemented by Python 3.7 and Tensorflow 2.3.0. Both the actor and critic networks of Alice and Bob have four connected hidden layers, each of which has 100 neurons and $elu(\cdot)$ is applied as the activation function. The activation function of the output layer of actor networks is $sigmoid(\cdot)$. The learning rate of actor networks is $10^{-4}$ and the learning rate of critic networks is $10^{-3}$. The AdamOptimizer~\cite{kingma2014adam} is applied. More parameters can be found in Table 1.

\begin{table}[!htbp]\label{tab2}
	\caption{Simulation Parameters}
	\begin{center}
		\begin{tabular}{|c|c|}
			\hline
			\textbf{Notation}&\textbf{Description} \\
			\hline
			M & 20\\
			\hline
			T & 19\\
            \hline
            $\gamma$ & 0.99\\
            \hline
            $n$ & 128\\
            \hline
            $N$ & 10000\\
            \hline
            $e^{max}$ & 3000\\
            \hline
            $\epsilon$ & 0.4\\
            \hline
            $\rho$ & 0.01\\
            \hline
            $E$ & 5\\
            \hline
            $\lambda$ & 0.5\\
            \hline
		\end{tabular}
	\end{center}
\end{table}

In order to evaluate the performance of the proposed FD2K scheme, we introduce the evaluation metrics for generated keys as follows:
\begin{itemize}
    \item \textbf{Key Agreement Rate (KAR)}: KAR represents the percentage of the same key bits between Alice and Bob, whose equation is described as a part of Eq.~\ref{reward_function}.
    \item \textbf{Randomness}: as the keys are generated for encryption/decryption algorithms, such as Advanced Encryption Standard (AES), and Data Encryption Standard (DES). It is widely acknowledged that the keys should be random to avoid force-brute attacks. National Institute of Standards and Technology (NIST)~\cite{8966} is proved to be an efficient random test suite to evaluate the randomness of both true and pseudo random number generators. Specifically, given the generated keys, each test of NIST will return a p-value. If the p-value is larger than the threshold, such as 0.01, we can confirm the generated keys are considered to pass the test.
    \item \textbf{Uniqueness}: uniqueness is defined to evaluate whether the physical signals can represent the different patterns and are different from each other. In this paper, the uniqueness is guaranteed by the different physical signals from Alice, Bob and Eve.
\end{itemize}

\begin{figure}[!htpb]
\centering
\includegraphics[scale=0.4]{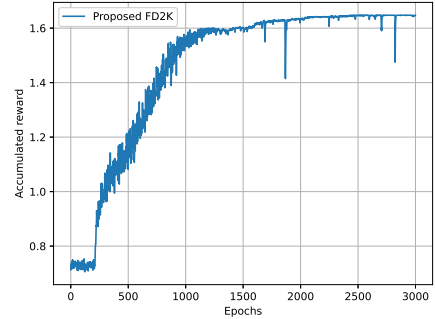}
\caption{Accumulated reward versus training epochs.}
\label{training_curve}
\end{figure}

First, we depict the accumulated reward of the proposed FD2K scheme in the training process in Fig.~\ref{training_curve}. Specifically, it is observed that the training curve remains below $0.8$ at the beginning. This is because the networks of the proposed FD2K are not updated and some poor attempts, which lead to lower rewards are explored by the agents of Alice and Bob. After that, the training curve starts to increase consistently and finally keeps stable above $1.6$, one plausible explanation is that the experience replay memory has enough transitions and  the training process starts. The networks of Alice and Bob are trained to capture the common features of physical signals between each other for key generation. 

\begin{figure}[!htpb]
\centering
\includegraphics[scale=0.4]{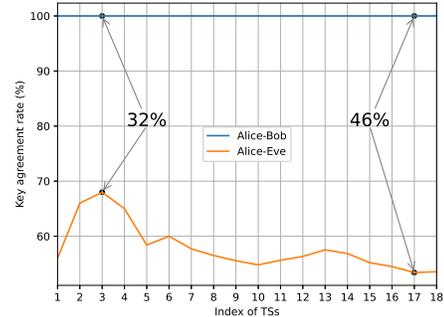}
\caption{Comparison of KAR of Alice-Bob and Alice-Eve.}
\label{key_compare}
\end{figure}

After the training stage, the networks of Alice and Bob are saved for evaluation. Note that our proposed FD2K scheme is based on the framework of centralized training and decentralized execution, both Alice and Bob enable to generate keys with their locally observed physical signals without any information/key exchange. As a result, the process of key generation can be fully distributed. We evaluate the performance of keys generated by FD2K in terms of KAR in Fig.~\ref{key_compare}. Considering the situation that most IoT devices embedded in networked systems in the real world are normally buried underground, the physical signals are extremely difficult to be obtained by the adversary, namely Eve, or easily to be found if inserting extra IoT devices. Additionally, in PLS, most of the digital attacks are based on the assumption that the IoT devices detached in the networked system are viewed as legitimate users, which means Eve can not insert any IoT device or can not obtain the perfect physical signals. Therefore, we consider a practical case that the actor network of Bob is attacked and known by Eve during the federated learning but it can not perfectly obtain the physical signals of Bob. Eve can capture the wireless signals from Alice and generate keys with different physical signals for information decryption. From Fig.~\ref{key_compare}, we can conclude that the average KAR of keys generated between Alice and Bob consistently outperforms the average KAR between Alice and Eve. Precisely, the generated keys between Alice and Bob have at least $32\%$ performance improvement when the index of TSs is 3 and have at most $46\%$ performance improvement when the index of TSs is 17, compared with the generated keys between Alice and Eve.

\begin{table}[!h]
\caption{p-values of NIST Test}
\centering
\begin{tabular}{|c|c|}
	\hline
	NIST Test & p-value \\
	\hline\hline
	Monobit Frequency & 0.184077 \\
	\hline
    Runs & 0.272715 \\
    \hline
    Discrete Fourier Transform & 0.520943 \\
    \hline
    Non Overlapping Template & 0.535998 \\
    \hline
    Approximate Entropy & 0.545495 \\
    \hline
    Cumulative Sums & 0.395994 \\
    \hline
    Random Excursion & 0.394067 \\
    \hline
    Random Excursion Variant & 0.600110 \\
    \hline
\end{tabular}
\label{p_value_compare}
\end{table}

Finally, we evaluate the randomness of keys generated by our proposed scheme in Table~\ref{p_value_compare} with Python implementation of the NIST test suite. It is clear to see that all p-values of keys are larger than $0.01$ and all generated keys pass the NIST test, which concludes that FD2K is efficient to extract common features of physical signals for key generation.

\section{Conclusion}\label{conclusion}
In this paper, we have proposed a distributed key generation scheme, which fully exploits the common features of physical dynamics embedded in networked systems for key generation. The proposed key generation scheme is based on multi-agent deep reinforcement learning approach, which enables legitimate users to generate keys in a distributed manner. Besides, the federated learning technique is applied to protect the privacy and security of local training data. Experimental results conclude that the proposed key generation scheme has efficient superiority for capturing common features between different physical signals and the generated keys have considerable performance improvement in terms of KAR, randomness, etc.

\section{Acknowledgement}

This work has been supported by the PETRAS National Centre of Excellence for IoT Systems Cybersecurity, which has been funded by the UK EPSRC under grant number EP/S035362/1.

\ifCLASSOPTIONcaptionsoff
  \newpage
\fi
\bibliographystyle{ieeetr}
\bibliography{reference}

\begin{thebibliography}{10}

\bibitem{256484}
U.~Maurer, ``Secret key agreement by public discussion from common
  information,'' {\em IEEE Transactions on Information Theory}, vol.~39, no.~3,
  pp.~733--742, 1993.

\bibitem{7809064}
J.~Zhang, B.~He, T.~Q. Duong, and R.~Woods, ``On the key generation from
  correlated wireless channels,'' {\em IEEE Communications Letters}, vol.~21,
  no.~4, pp.~961--964, 2017.

\bibitem{10.1145/1409944.1409960}
S.~Mathur, W.~Trappe, N.~Mandayam, C.~Ye, and A.~Reznik, ``Radio-telepathy:
  Extracting a secret key from an unauthenticated wireless channel,'' in {\em
  Proceedings of the 14th ACM International Conference on Mobile Computing and
  Networking}, MobiCom '08, (New York, NY, USA), p.~128–139, Association for
  Computing Machinery, 2008.

\bibitem{6226870}
Y.~Liu, S.~C. Draper, and A.~M. Sayeed, ``Exploiting channel diversity in
  secret key generation from multipath fading randomness,'' {\em IEEE
  Transactions on Information Forensics and Security}, vol.~7, no.~5,
  pp.~1484--1497, 2012.

\bibitem{halperin2011tool}
D.~Halperin, W.~Hu, A.~Sheth, and D.~Wetherall, ``Tool release: Gathering
  802.11 n traces with channel state information,'' {\em ACM SIGCOMM computer
  communication review}, vol.~41, no.~1, pp.~53--53, 2011.

\bibitem{7063613}
J.~Zhang, A.~Marshall, R.~Woods, and T.~Q. Duong, ``Secure key generation from
  ofdm subcarriers' channel responses,'' in {\em 2014 IEEE Globecom Workshops
  (GC Wkshps)}, pp.~1302--1307, 2014.

\bibitem{6914340}
W.~Xi, X.-Y. Li, C.~Qian, J.~Han, S.~Tang, J.~Zhao, and K.~Zhao, ``Keep: Fast
  secret key extraction protocol for d2d communication,'' in {\em 2014 IEEE
  22nd International Symposium of Quality of Service (IWQoS)}, pp.~350--359,
  2014.

\bibitem{ruotsalainen2019experimental}
H.~Ruotsalainen, J.~Zhang, and S.~Grebeniuk, ``Experimental investigation on
  wireless key generation for low-power wide-area networks,'' {\em IEEE
  Internet of Things Journal}, vol.~7, no.~3, pp.~1745--1755, 2019.

\bibitem{8519327}
J.~Zhang, A.~Marshall, and L.~Hanzo, ``Channel-envelope differencing eliminates
  secret key correlation: Lora-based key generation in low power wide area
  networks,'' {\em IEEE Transactions on Vehicular Technology}, vol.~67, no.~12,
  pp.~12462--12466, 2018.

\bibitem{8433175}
L.~Jiao, J.~Tang, and K.~Zeng, ``Physical layer key generation using virtual
  aoa and aod of mmwave massive mimo channel,'' in {\em 2018 IEEE Conference on
  Communications and Network Security (CNS)}, pp.~1--9, 2018.

\bibitem{zhang2021h2k}
J.~Zhang, Y.~Zheng, W.~Xu, and Y.~Chen, ``H2k: A heartbeat-based key generation
  framework for ecg and ppg signals,'' {\em IEEE Transactions on Mobile
  Computing}, 2021.

\bibitem{7944621}
X.~Fang, N.~Zhang, S.~Zhang, D.~Chen, X.~Sha, and X.~Shen, ``On physical layer
  security: Weighted fractional fourier transform based user cooperation,''
  {\em IEEE Transactions on Wireless Communications}, vol.~16, no.~8,
  pp.~5498--5510, 2017.

\bibitem{s22103951}
Z.~Wei, L.~Wang, S.~C. Sun, B.~Li, and W.~Guo, ``Graph layer security:
  Encrypting information via common networked physics,'' {\em Sensors},
  vol.~22, no.~10, 2022.

\bibitem{yang2019federated}
Q.~Yang, Y.~Liu, T.~Chen, and Y.~Tong, ``Federated machine learning: Concept
  and applications,'' {\em ACM Transactions on Intelligent Systems and
  Technology (TIST)}, vol.~10, no.~2, pp.~1--19, 2019.

\bibitem{mnih2015human}
V.~Mnih, K.~Kavukcuoglu, D.~Silver, A.~A. Rusu, J.~Veness, M.~G. Bellemare,
  A.~Graves, M.~Riedmiller, A.~K. Fidjeland, G.~Ostrovski, {\em et~al.},
  ``Human-level control through deep reinforcement learning,'' {\em nature},
  vol.~518, no.~7540, pp.~529--533, 2015.

\bibitem{10.5555/3009657.3009806}
R.~S. Sutton, D.~McAllester, S.~Singh, and Y.~Mansour, ``Policy gradient
  methods for reinforcement learning with function approximation,'' in {\em
  Proceedings of the 12th International Conference on Neural Information
  Processing Systems}, NIPS'99, (Cambridge, MA, USA), p.~1057–1063, MIT
  Press, 1999.

\bibitem{lowe2017multi}
R.~Lowe, Y.~Wu, A.~Tamar, J.~Harb, P.~Abbeel, and I.~Mordatch, ``Multi-agent
  actor-critic for mixed cooperative-competitive environments,'' {\em Neural
  Information Processing Systems (NIPS)}, 2017.

\bibitem{mcmahan2017communication}
B.~McMahan, E.~Moore, D.~Ramage, S.~Hampson, and B.~A. y~Arcas,
  ``Communication-efficient learning of deep networks from decentralized
  data,'' in {\em Artificial intelligence and statistics}, pp.~1273--1282,
  PMLR, 2017.

\bibitem{rossman2000epanet}
L.~A. Rossman {\em et~al.}, ``{EPANET} 2: users manual,'' 2000.

\bibitem{kingma2014adam}
D.~P. Kingma and J.~Ba, ``Adam: A method for stochastic optimization,'' {\em
  arXiv preprint arXiv:1412.6980}, 2014.

\bibitem{8966}
L.~Bassham, A.~Rukhin, J.~Soto, J.~Nechvatal, M.~Smid, S.~Leigh, M.~Levenson,
  M.~Vangel, N.~Heckert, and D.~Banks, ``A statistical test suite for random
  and pseudorandom number generators for cryptographic applications,''
  2010-09-16 2010.

\end{thebibliography}
\end{document}